\documentclass{gOMS2e}
\usepackage{tipa}
\usepackage{amssymb,amsmath,amsthm}
\usepackage{subfigure}% Support for small, `sub' figures and tables
\usepackage{graphicx}
\usepackage{epstopdf}
\usepackage{breqn}
\usepackage[usenames, dvipsnames]{color}

\usepackage{epstopdf}% To incorporate .eps illustrations using PDFLaTeX, etc.

\theoremstyle{plain}% Theorem-like structures

\theoremstyle{definition}

\theoremstyle{remark}

\begin{document}

%\jvol{00} \jnum{00} \jyear{2015} \jmonth{July}

%\articletype{GUIDE}

\title{{\itshape Speech Map: A Statistical Multimodal Atlas of 4D Tongue Motion During Speech from Tagged and Cine MR Images}}

\author{Jonghye Woo$^{\rm a}$$^{\ast}$\thanks{$^\ast$Email: jwoo@mgh.harvard.edu. Jonghye Woo and Fangxu Xing contributed equally to this work.
\vspace{6pt}}, Fangxu Xing$^{\rm a}$, Maureen Stone$^{\rm b}$, Jordan Green$^{\rm c}$, Timothy G. Reese$^{\rm d}$, Thomas J. Brady$^{\rm a}$, Van J. Wedeen$^{\rm d}$, Jerry L. Prince$^{\rm e}$, and Georges El Fakhri$^{\rm a}$\\\vspace{6pt}$^{a}${\em{Gordon Center for Medical Imaging, Department of Radiology, Massachusetts General Hospital, Harvard Medical School, Boston, MA 02114, USA}};$^{b}${\em{Department of Neural and Pain Sciences, University of Maryland Dental School, Baltimore, MD 21201, USA}}; $^{c}${\em{Department of Communication Sciences and Disorders, MGH Institute of Health Professions, Boston, MA 02129, USA}}; $^{d}${\em{Athinoula A. Martinos Center for Biomedical Imaging, Department of Radiology, Massachusetts General Hospital, Harvard Medical School, Boston, MA 02129, USA}}; $^{e}${\em{Department of Electrical and Computer Engineering, Johns Hopkins University, Baltimore, MD 21218, USA}}}

\maketitle

\begin{abstract}

Quantitative measurement of functional and anatomical traits of 4D tongue motion in the course of speech or other lingual behaviors remains a major challenge in scientific research and clinical applications. Here, we introduce a statistical multimodal atlas of 4D tongue motion using healthy subjects, which enables a combined quantitative characterization of tongue motion in a reference anatomical configuration. This atlas framework, termed Speech Map, combines cine- and tagged-MRI in order to provide both the anatomic reference and motion information during speech. Our approach involves a series of steps including (1) construction of a common reference anatomical configuration from cine-MRI, (2) motion estimation from tagged-MRI, (3) transformation of the motion estimations to the reference anatomical configuration, and (4) computation of motion quantities such as Lagrangian strain. Using this framework, the anatomic configuration of the tongue appears motionless, while the motion fields and associated strain measurements change over the time course of speech. In addition, to form a succinct representation of the high-dimensional and complex motion fields, principal component analysis is carried out to characterize the central tendencies and variations of motion fields of our speech tasks. Our proposed method provides a platform to quantitatively and objectively explain the differences and variability of tongue motion by illuminating internal motion and strain that have so far been intractable. The findings are used to understand how tongue function for speech is limited by abnormal internal motion and strain in glossectomy patients.

\end{abstract}

\section{Introduction}

 The human tongue is a muscular hydrostat~\cite{kier1985} and is considered to be a complex biomechanical system comprised of numerous intrinsic and extrinsic muscles. In the course of speech or other lingual behaviors, the human tongue takes on a variety of positions, shapes, and local deformations created by the complex interactions of its inter-digitated muscles~\cite{gilbert2006}. Understanding the interactions of these components is crucial for many applications, such as speech production or swallowing, and often requires anatomical and motion information ranging from voxel level to muscle level; integrative models of tongue anatomy and physiology are important for understanding the mechanisms of speech production as well as disease and planning intervention. Visualization and quantification of tongue motion during speech or swallowing through medical imaging such as ultrasound or magnetic resonance imaging (MRI) have been used for past decades, yet combined quantitative measurement of functional and anatomical features in a common reference space still remains an unmet goal. 

\def\FigureHeight{110mm}
\begin{figure}[htb]
 \center{
 \begin{tabular}{c@{ }c}
   \includegraphics[trim=0mm 0mm 0mm
0mm,clip=true,height=\FigureHeight]{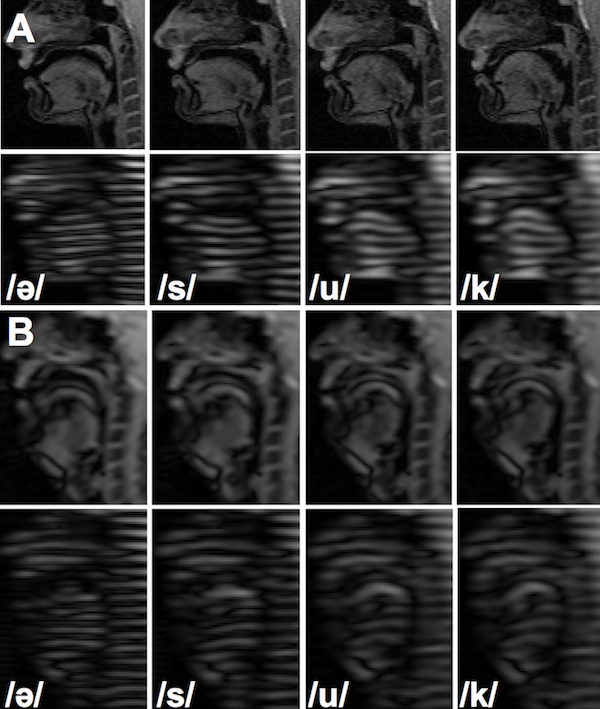} \\
\end{tabular}}
\caption{Example of speech data (``\textschwa-suk'') from a healthy control in (a) and a glossectomy patient in (b) at four representative time frames. Cine- and tagged-MRI are shown in the first and second rows, respectively. Note that both modalities are in the same spatio-temporal coordinate space.}\label{fig:example}
\end{figure}

Numerous studies have been conducted to understand the function of the tongue to date. Research interest in tongue motion during speech or swallowing using MRI has grown in recent years as technology has the potential to elucidate mechanisms of tongue motion related disorders. In particular, it is critical to obtain the 4D (3D space with time) motion information about speech movements to understand and model the speech production process. Multimodal MR imaging and subsequent image analysis has allowed us to investigate the multifaceted nature of tongue structure and function. For instance, structural MRI and diffusion MRI~\cite{Gaige2007, Shinagawa2008} provide an exquisite depiction of internal muscular architecture and local fiber orientations, respectively. In addition, real-time MRI provides the ability to examine real-time changes in the vocal tract shaping in speech production~\cite{Naranayan2004,fu2016speech}. Furthermore, tagged-MRI~\cite{Parthasarathy2007,pipeline} has allowed us to track the internal motion of the tongue in addition to motion on its surface. For example, Fig.~\ref{fig:example} illustrates an example of speech data from cine- and tagged-MRI for a normal control and a glossectomy patient, showing different motion patterns at different time points during speech. However, it is challenging to contrast and compare the differences between the two subjects as visually assessed. Therefore, the development of meaningful quantitative measurements from these imaging techniques facilitates quantitative comparison across controls and diseased populations. In addition, the usefulness of these methodologies is dependent on the development of meaningful spatio-temporal indices of tongue function and of techniques for integrating complementary information across these imaging modalities.

Although MRI has played a pivotal role in tongue image and motion analysis, advancement in speech science research is hampered by the lack of tools for accurate, reproducible,
and automated quantitative characterization of tongue motion during speech in a common reference space in order to examine tongue motion and its variability. This is partly because the size, shape, or motion pattern of the tongue during speech may vary from one subject to another, yet there is no comprehensive and systematic framework to examine the difference and variability in a common reference space. For instance, most of the work using tagged-MRI have been carried out to analyze subject-specific internal tongue motion patterns and therefore it is difficult to objectively and quantitatively compare different motion patterns, especially patient motion patterns. Despite its potential for analyzing internal tongue motion data in an objective manner, there has been very little research toward an atlas of 4D tongue motion during speech. Two key works related to the development of such a 4D atlas are \cite{motionless} and \cite{woo_cmbbe_2016}. Wedeen et al.~\cite{motionless} introduced a new data acquisition strategy within a Lagrangian framework to analyze myocardial strain-rates of the heart. That work was performed on a subject-by-subject basis to analyze motion patterns using strain rates. Woo et al.~\cite{woo_cmbbe_2016} proposed a 4D atlas of the tongue during speech from cine-MRI using diffeomorphic groupwise registration. In that work, only cine-MRI was used to generate the average motion pattern of the tongue during speech. The present work is motivated by both approaches in the sense that the average speech movements using healthy speakers are constructed in a single anatomical configuration, the common reference space. We use both cine- and tagged-MRI to characterize motion quantities during speech in an objective and quantitative manner. The Speech Map uses a fixed anatomical reference space from cine-MRI which characterizes speech patterns from tagged-MRI. The way it works is that tagged-MRI from a specific speaker are overlaid to an atlas reference space that provides vocal tract tissue motion patterns and structural boundaries. The individual speaker's motion patterns are then registered and analyzed in the space, thus allowing comparisons between subjects with different properties such as motion fields and strains.

In this work, we present a novel approach to combining imaging information by providing a sequence of images in which the tongue anatomical configuration appears frozen~\cite{motionless}, but in which each voxel location displays the displacement and associated motion quantities of a fixed tissue element in a common reference space. This approach allows us to quantitatively assess differences in speech movements across speakers and speech-impaired populations. In contrast to the previous methods of tongue imaging in spatial coordinates~\cite{woo_cmbbe_2016,eulerian_atlas}, our approach provides movies of the tongue in material coordinates. In addition, each subject's tongue motion is transformed to the common anatomical configuration, thereby allowing us to objectively compare patients to the atlas and to each other. To the best of our knowledge, this is the first attempt at constructing a statistical multimodal atlas of 4D tongue motion during speech using both cine- and tagged-MRI within a Lagrangian framework. Using this framework, we demonstrate that accurate characterization of 4D tongue motion during speech is possible, thereby establishing the normal motion patterns and associated quantitative measures in a reference configuration. We also provide data demonstrating the application of the technique for understanding the mechanisms of tongue impairment in an individual's abnormal speech due to glossectomy.

The remainder of this paper is structured as follows. In Section 2, prior work on 4D multimodal atlas construction is reviewed. The atlas building method for the tongue during speech is presented in Section 3. In Section 4, we describe experimental results. A detailed discussion is presented in Section 5 and finally conclusions and future directions are given in Section 6.

\section{Related Work}

In this section, we review multimodal 4D atlas construction methods. 4D atlas construction has been an area of active research in recent years. The ability to construct a representative 4D atlas of a population is an important tool in the analysis and interpretation of medical images in organs such as the heart, the fetal brain, the tongue, and the lung. 4D atlases provide changes in anatomical references or particular features of an object over time. 4D atlases become 4D statistical or probabilistic atlases when they represent the differences within a population of subjects~\cite{woo_cmbbe_2016}. 4D atlases have numerous applications in medical image analysis. For example, 4D atlases representing normal growth or motion patterns can be used to detect abnormalities or potential disease by measuring the variation of a subject relative to the variations contained in the atlas. In addition, they can also provide an a priori information for the segmentation~\cite{segmentation} and registration of anatomical structures. Compared with static 3D multimodal atlas construction methods or 4D atlas construction methods that use a single modality, the multimodal 4D atlas construction problem is challenging as both nonlinear mappings in both different modalities and motion modeling need to be performed cooperatively, which may be in a sequential or joint manner. 

Several approaches have been proposed to construct such 4D atlases. Xing et al.~\cite{eulerian_atlas} proposed a 4D multimodal atlas of the tongue during speech within an Eulerian framework. In the present work, we also used both cine- and tagged-MRI, but adopted a Lagrangian framework, where the reference anatomic configuration remains fixed, while motion fields change in the course of speech. Puyol-Anton et al.~\cite{motion_atlas_heart} presented a multimodal cardiac motion atlas construction method, in which both MRI and ultrasound were used to construct the atlas. In that work, high-quality tagged-MRI data were first used to form an atlas, and patient data from ultrasound were related to the tagged-MRI based atlas. The approach for embedding the displacement vector was based on principal component analysis (PCA), which could be improved by using nonlinear manifold learning methods. Furthermore, a 4D statistical atlas construction method was presented to build a swine heart atlas from PET-CT images~\cite{motion_atlas_swine_heart}. In that work, the data were acquired from a hybrid PET-CT scanner and spatially co-registered PET and CT data were assumed. A hierarchical normalization method was then used to progressively construct the atlas from anatomic images (i.e., CT angiography) to functional images (i.e., PET). In the present work, both cine- and tagged-MRI are in the same spatio-temporal coordinate system and therefore the nonlinear mappings learned in one modality can be used to map the other modality. In related developments, Wang et al.~\cite{wang2016} proposed a joint segmentation and registration method to model 4D changes in pathological anatomy across time by providing an explicit mapping of a healthy normative template. In that work, because a normative template cannot deal with pathological appearance for the joint segmentation and registration, they used different options for initialization via a supervised and semi-supervised learning and transfer learning approach for the application of traumatic brain injury.

\section{Materials and Methods}
\subsection{Data Acquisition}
\subsubsection{Data Collection and Preprocessing}

In our study, speech MRI data were collected from fourteen healthy subjects and two glossectomy patients (native speakers). Subjects repeated a pre-trained speech task (i.e., ``\textschwa-suk''), where the speech task lasts for 1 second, when cine- and tagged-MR images were acquired as a sequence of image frames at multiple parallel slice locations that cover a region of interest encompassing the tongue and the surrounding structures. We used a segmented k-space data acquisition scheme, which is not real-time but requires a number of repetitions~\cite{Mcveigh_1992} to obtain a good representation of tongue motion associated with the speech task. We used T2-weighted multi-slice 2D dynamic cine- and tagged-MRI data at a frame rate of 26 frames per second using a Siemens 3.0 T Tim Treo system (Siemens Medical Solutions, Malvern, PA) with 12-channel head and 4-channel neck coil. To avoid the blurred effect caused by involuntary motion such as swallowing, three orthogonal stacks from axial, coronal, and sagittal orientations were acquired to cover the whole tongue. Each dataset had 6 mm slice thickness and 1.8 mm in-plane resolution. Other sequence parameters included repetition time (TR) 36 ms, echo time (TE) 1.47 ms, flip angle 6$^{\circ}$, and turbo factor 11. Super-resolution volume reconstruction was then used to create a single volume by combining all three stacks with an isotropic resolution~\cite{supres}. For tagged-MRI, the acquisition matrix was the same as cine-MRI and complementary spatial modulation of magnetization (CSPAMM) was applied. The datasets had 26 frames per second with a temporal resolution of 36 ms for each phase with no delay from the tagging pulse, 6 mm thick slices (12 mm sinusoidal tag period), and 1.875 mm in-plane resolution with no gap. The field-of-view was 24 cm. Note that both cine- and tagged-MRI were in the same spatio-temporal coordinate system.

\subsubsection{Subjects and Speech task}
The atlas is constructed from a database of cine- and tagged-MR images of fourteen healthy native subjects. The sample population includes both males and females with ages ranging from 21 to 57. Table~\ref{table:subject} lists detailed information on age, weight, and gender included in the atlas construction. The MRI speech task is the phrase ``\textschwa-suk.'' The word begins with a neutral tongue position (schwa). The tongue body motion is simple because it moves only forward or backward, and the phrase uses little to no jaw motion, thus increasing tongue deformation. There are four distinctive time frames /\textschwa/, /s/, /u/, and /k/ in that phrase. 

%% Table 1 %%%%%%%%%%%%%%%%%%%%%%%%%%%%%%%%%%%%%%%%%%%%%%%%%%%%%%%
\begin{table}[h]
\centering
\caption{Detailed characteristics of the fourteen healthy subjects}
\begin{tabular}{c|c|c|c||c|c|c|c} \hline
 Subjects & Age & Gender & Weight (lb) & Subjects & Age & Gender & Weight (lb)  \\
 \hline \hline
 1 & 23 & M & 155 & 8 & 21 & F & 126 \\
 2 & 31 & F & 150 & 9 & 37 & M & 150 \\
 3 & 24 & F & 100 & 10 & 22 & M & 130 \\
 4 & 57 & F & 170 & 11 & 43 & M & 180 \\
 5 & 43 & F & 217 & 12 & 26 & M & 240 \\
 6 & 35 & M & 210 & 13 & 42 & F & 180 \\
 7 & 45 & F & 180 & 14 & 52 & M & 156 \\
 
   \hline 
\end{tabular}\label{table:subject}
\end{table}
%%%%%%%%%%%%%%%%%

\def\FigureHeight{15mm}
\begin{figure}[htb]
 \center{
 \begin{tabular}{c@{ }c}
   \includegraphics[trim=0mm 0mm 0mm
0mm,clip=true,height=\FigureHeight]{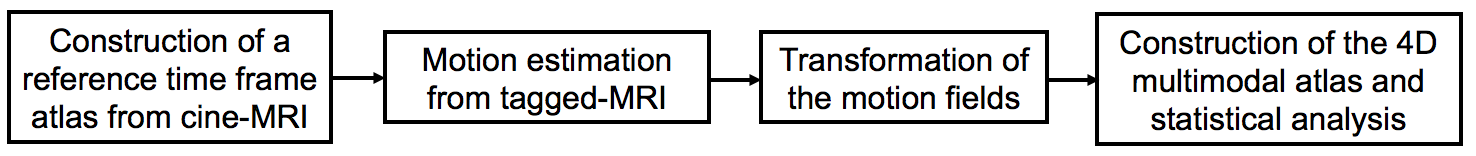} \\
\end{tabular}}
\caption{Flowchart of the proposed Speech Map}\label{fig:flowchart}
\end{figure}

\subsection{4D Multimodal Atlas Construction Method}
This section introduces our method to construct the 4D multimodal atlas using both cine- and tagged-MRI. Our approach involves a series of steps illustrated in Fig.~\ref{fig:flowchart}.

\subsubsection{Construction of Reference Atlas from Cine-MRI}
The first step is to create an atlas of the reference time frame (i.e., time frame 1) from cine-MRI, which serves as a reference anatomical configuration. For $n$ images, given a set of points in the common coordinates of the tongue in the undeformed material time frame, $\Omega_r \subset \mathbb{R}^3 \to \mathbb{R}$, and a set of points in each image space $\Omega_i\subset \mathbb{R}^3 \to \mathbb{R}$, $i=1, \cdot \cdot \cdot, n$, the goal of the atlas construction is to find a set of diffeomorphic mappings $M$, each of which transforms any point in each individual image space to a corresponding point in image $I_i$: $M = [{{\phi_i}:{\mathbf{x}_i} \mapsto {\mathbf{x}_r}, i = 1, \cdots ,n}$]. We use an unbiased groupwise diffeomorphic registration approach with a cross-correlation similarity metric~\cite{ants}. The atlas creation procedure combines a groupwise affine registration as an initial transformation, followed by a groupwise deformable registration. We then get both forward (i.e., $\phi_i$) and inverse (i.e.,$\phi_i^{-1}$) mappings during the atlas building process.

\subsubsection{Motion Estimation from Tagged-MRI}
The phase vector incompressible registration algorithm (PVIRA)~\cite{pvira} is used for 3D motion estimation from tagged-MRI. In brief, PVIRA computes a dense 3D motion field at each time frame via a series of steps described as follows. It interpolates 2D tagged slices into 3D volumes and then uses a harmonic phase filter~\cite{harp} to produce a series of phase volumes from interpolated 3D volumes. The phase volume pairs from axial, sagittal, and coronal directions are then processed via an incompressible iterative image registration framework~\cite{ilogdemons}. We denote these phase volumes by $\Phi_{a0}$, $\Phi_{at}$, $\Phi_{s0}$, $\Phi_{st}$, $\Phi_{c0}$, and $\Phi_{ct}$, where $a$, $s$, and $c$ stand for axial, sagittal, and coronal tag directions and the two time frames are $0$ and $t$. The symmetric  velocity field update of PVIRA, defined at each voxel of the three volumes, is defined as:
\begin{equation}\label{eqnewdemonup}
\begin{aligned}
\delta \mathbf{v}(\mathbf{x}) &= \frac{\mathbf{v}_0(\mathbf{x})}{\alpha_1(\mathbf{x}) + \alpha_2(\mathbf{x})/K},
\end{aligned}
\end{equation}
where $K$ is the normalization factor and $\mathbf{v}_0(\mathbf{x})$, $\alpha_1(\mathbf{x})$, and $\alpha_2(\mathbf{x})$ are given by
\begin{equation}\label{eqnewdemonup}
\begin{aligned}
\mathbf{v}_0(\mathbf{x}) &= W(\Phi_{a0}(\mathbf{x}) - \Phi_{at}(\mathbf{x}))(\nabla^* \Phi_{a0}(\mathbf{x}) + \nabla^* \Phi_{at}(\mathbf{x})) \\
&+ W(\Phi_{s0}(\mathbf{x}) - \Phi_{st}(\mathbf{x}))(\nabla^* \Phi_{s0}(\mathbf{x}) + \nabla^* \Phi_{st}(\mathbf{x})) \\
&+ W(\Phi_{c0}(\mathbf{x}) - \Phi_{ct}(\mathbf{x}))(\nabla^* \Phi_{c0}(\mathbf{x}) + \nabla^* \Phi_{ct}(\mathbf{x})) \;, \\
\alpha_1(\mathbf{x}) &= ||\nabla^* \Phi_{a0}(\mathbf{x}) + \nabla^* \Phi_{at}(\mathbf{x})||^2 + ||\nabla^* \Phi_{s0}(\mathbf{x}) + \nabla^* \Phi_{st}(\mathbf{x})||^2 \\
&+ ||\nabla^* \Phi_{c0}(\mathbf{x}) + \nabla^* \Phi_{ct}(\mathbf{x})||^2 \;, \\
\alpha_2(\mathbf{x}) &= W(\Phi_{a0}(\mathbf{x}) - \Phi_{at}(\mathbf{x}))^2 + W(\Phi_{s0}(\mathbf{x}) - \Phi_{st}(\mathbf{x}))^2 \\
&+ W(\Phi_{c0}(\mathbf{x}) - \Phi_{ct}(\mathbf{x}))^2 \;.
\end{aligned}
\end{equation}
Note that a wrapping operator $W(\theta)$ and the modified gradient operator used here are given by 
\begin{equation}\label{eqphasegrad}
W(\theta) = \mathrm{mod} (\theta + \pi, 2\pi) - \pi
\end{equation}
and 

\begin{equation}\label{eqphasegrad}
\nabla^*\Phi(\mathbf{x}) :=
\left\{
\begin{array}{l}
\nabla\Phi(\mathbf{x}), \qquad \qquad \mathrm{if} \; |\nabla\Phi(\mathbf{x})| \leq |\nabla W(\Phi(\mathbf{x})+\pi)| \;,\\
\nabla W(\Phi(\mathbf{x})+\pi),  \mathrm{otherwise} \\
\end{array}
\right.
\end{equation}

The forward and inverse deformation fields that are incompressible and diffeomorphic are given by
\begin{equation}\label{eqphasegrad_}
\varphi(\mathbf{x}) = \exp (\mathbf{v}(\mathbf{x}))\,\,\,\mathrm{and}\,\, \varphi^{-1}(\mathbf{x}) = \exp(-\mathbf{v}(\mathbf{x})).
\end{equation}

In PVIRA, incompressibility is only enforced where tissue is defined by using a combined HARP magnitude image, which is computed from the three HARP magnitude images that are found at the same time as the HARP phase images. An inverse motion field is produced along with the forward motion field, enabling both an Eulerian and a Lagrangian output between the undeformed static tongue and its deformed state. For a full description of the motion estimation pipeline, the readers can refer to the work~\cite{pipeline}.

\def\FigureHeight{80mm}
\begin{figure}[htb]
 \center{
 \begin{tabular}{c@{ }c}
   \includegraphics[trim=0mm 0mm 0mm
0mm,clip=true,height=\FigureHeight]{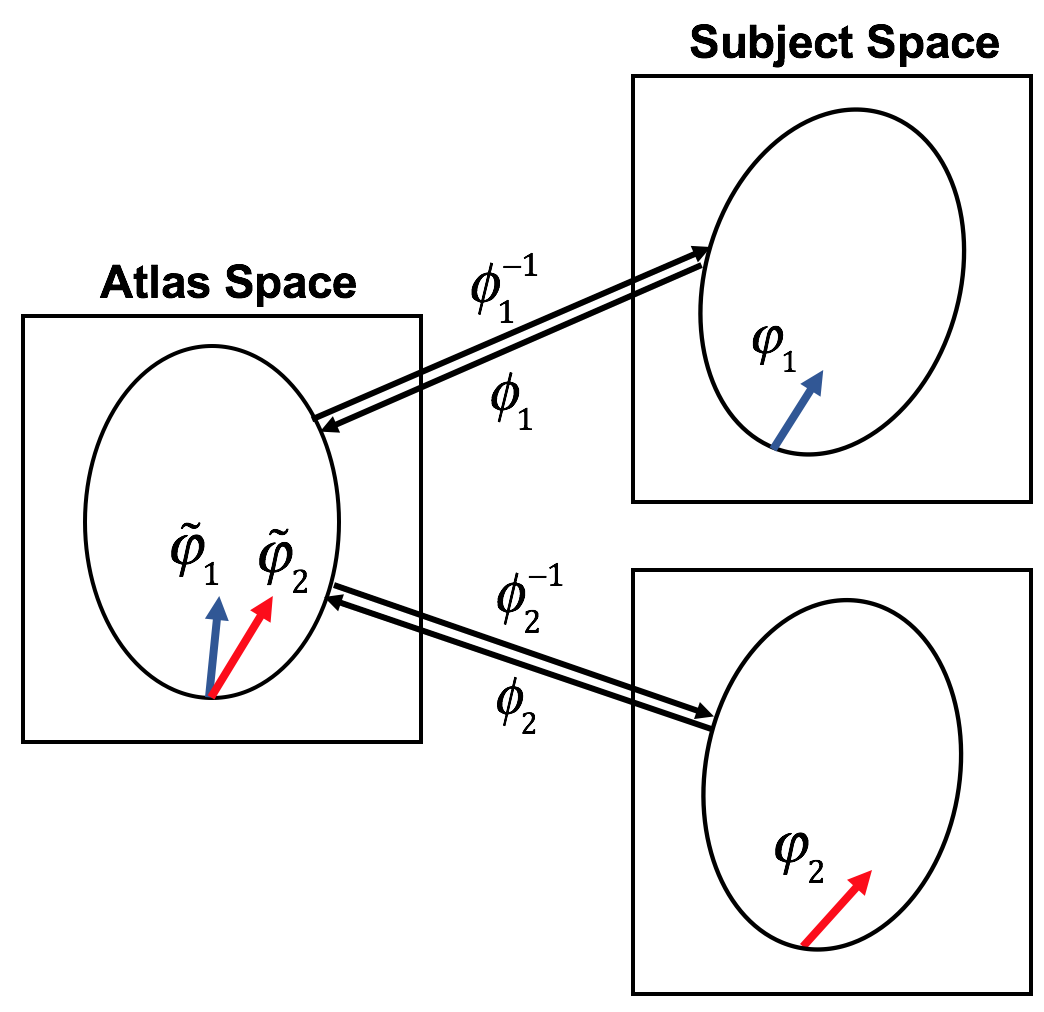} \\
\end{tabular}}
\caption{Schematic illustration of the transformation of the motion fields derived in the subject space to the atlas space.}\label{fig:xform}
\end{figure}
%%%%%%%%%%%%%%%%%%%%%%%%%%%%%%%%%%%%%%%%%%%%%%%%%%%%%%%%%%%%%%%%%%%%%

\subsubsection{Transformation of Motion Estimation}

The PVIRA motion fields derived from tagged-MRI are diffeomorphic mappings defined in the coordinate system of subject $i$: $\varphi_i: \Omega_s \to \Omega_s, i = 1, \cdots, n$. The transformation of the subject-specific motion fields into the atlas coordinate system is given by~\cite{ct_lung}

\begin{equation}\label{eq:combination}
\tilde \varphi_i(\mathbf{x}) = {\phi_i} \circ \varphi_i \circ \phi_i^{-1}(\mathbf{x}), \mathbf{x} \in D
\end{equation}
where $D \subset \Omega_r$ is a subset of pixels including the tongue region in the atlas space and $\tilde \varphi_i$ denotes the transformed motion field of subject $i, i = 1, \cdots, n$. Fig.~\ref{fig:xform} illustrates this transformation process, where two subject-specific diffeomorphic motion fields indicated by blue and red arrows are transformed using the diffeomorphic deformation fields learned during the reference atlas construction process.

\subsubsection{Computation of Motion Quantities}
Lagrangian strain tensors are computed from PVIRA in the subject coordinate system or atlas coordinate system to reflect the tongue tissue's local deformation at every time frame. We denote the estimated motion field by $\mathbf{u}(\mathbf{X})$, where $\mathbf{X}$ is the coordinates of the tongue in the undeformed material time frame. The deformation gradient tensor $\mathbf{F}(\mathbf{X})$ can be computed as $\mathbf{F}(\mathbf{X}) = \mathbf{I}+d\mathbf{u}/d\mathbf{X}$. The Lagrangian strain tensor ${\bf E}$ is defined as
\begin{equation}
\mathbf{E}(\mathbf{X}) = \frac{1}{2}(\mathbf{F}(\mathbf{X})^T\mathbf{F}(\mathbf{X})-\mathbf{I}),
\end{equation}
which used in the undeformed material frame to evaluate how much a given displacement
differs locally from a rigid body displacement. The eigen-decomposition of the Lagrangian strain yields three principal directions and values (E1, E2, and E3; E1 $\ge$ E2 $\ge$ E3) that indicate the directions and amount of the major extension, the second major deformation (either stretching or compression), and the major compression, respectively. In addition, we use the mean of the magnitude of the motion field, which is defined as
\begin{equation}\label{eq:mean_deform}
\mathbf{MD} = \frac{{\int_{{\Omega}} {\left| \mathbf{u}(\mathbf{X}) \right| B_S(\mathbf{X})d\mathbf{X}}}}{{\int_{{\Omega}} {B_S(\mathbf{X})d\mathbf{X}}}},
\end{equation}
where $\left|\cdot\right|$ denotes the magnitude of a motion field and $B_S(\cdot)$ is the bounding box encompassing the tongue region, which is defined in the reference time frame as
\begin{equation}\label{eq:character}
{B_S}(\mathbf{X}) = \left\{ \begin{array}{l}
1\,\,\mathrm{if}\,\mathbf{X} \in S\\
0\,\,\mathrm{if}\,\mathbf{X} \notin S
\end{array} \right.
\end{equation}

\subsubsection{Statistical Analysis Using PCA}
We perform PCA on the Lagrangian motion fields in the four distinctive time frames, /\textschwa/, /s/, /u/, and /k/, across subjects. The goal of our statistical motion model using PCA is to form a succinct representation of the high-dimensional motion data by building a model of a class of internal motion patterns given a set of examples of the internal motion patterns during speech. All computations are performed in the atlas coordinate system. The procedure of PCA is to approximate matrix $\mathbf{C}$ using a linear model of the form similar to~\cite{woo_cmbbe_2016}:
\begin{equation}\label{eq:sdm}
\mathbf{C} =  \mathbf{\hat C} + \Phi \mathbf{b}\,
\end{equation}
where $\mathbf{\hat C}$ is the mean of the Lagrangian motion fields for all subjects 
\begin{equation}\label{eq:mean_deform}
\mathbf{\hat C} = \frac{1}{n}\sum\limits_{i = 1}^{n} \tilde \varphi_i,
\end{equation}
$\mathbf{b}$ is the parameter vector, and $\tilde \varphi_i$ is the Lagrangian motion field of subject $i$ in the atlas space. The columns of the matrix $\Phi$ are formed by the principal components (PC) of the covariance matrix $\mathbf{S}$
\begin{equation}\label{eq:cov}
\mathbf{S} = \frac{1}{{n - 1}}\sum\limits_{i = 1}^n {({\tilde \varphi_i} - \mathbf{\hat C}){{({\tilde \varphi_i} - \mathbf{\hat C})}^T}}.
\end{equation}

\section{Experiments and Results}

In this section, we present results of experiments on the \emph{in vivo} tongue data, which demonstrates the performance of the proposed method. All programs were implemented using either C++ or MATLAB. In our experiments, we primarily focused on the analysis of the four representative time frames of the sounds /\textschwa/, /s/, /u/, and /k/.

\def\FigureHeight{32mm}
\begin{figure}[htb]
 \center{
 \begin{tabular}{c@{ }c}
   \includegraphics[trim=0mm 0mm 0mm
0mm,clip=true,height=\FigureHeight]{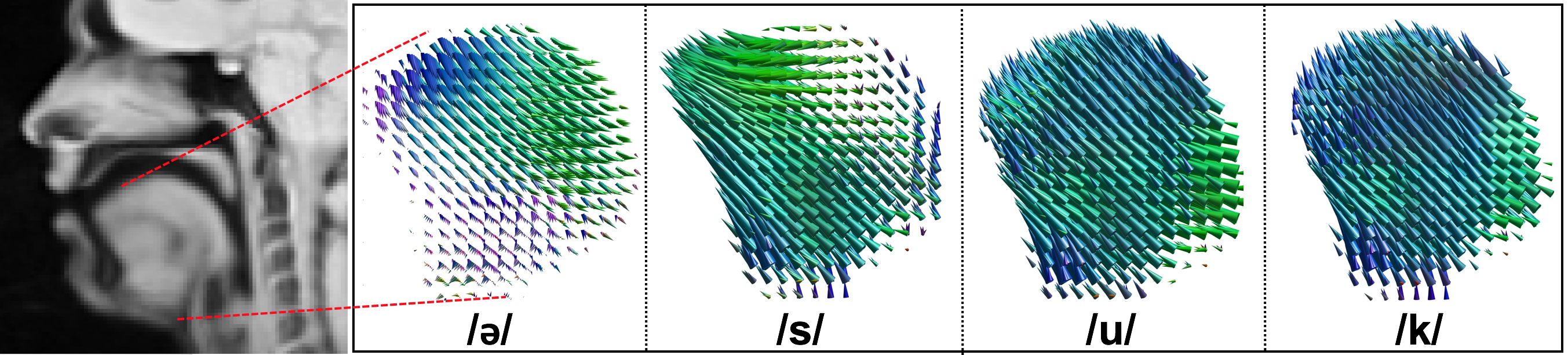} \\
\end{tabular}}
\caption{Plot of the reference atlas using the first time frame from cine-MRI and our final atlas motion fields from ``\textschwa-suk.'' Note that motion fields are rooted in the material coordinates in the reference atlas space.}\label{fig:final_motion}
\end{figure}
%%%%%%%%%%%%%%%%%%%%%%%%%%%%%%%%%%%%%%%%%%%%%%%%%%%%%%%%%%%%%%%%%%%%%

Fig.~\ref{fig:final_motion} depicts the reference atlas constructed using cine-MRI, which serves as a reference anatomical configuration, and average motion fields of ``\textschwa-suk.'' The motion fields of each time frame are not biased by any specific subject's anatomic features, thereby allowing us to characterize and compare the motion fields directly at each voxel. Each time frame describes representative motion fields of the classes of each phoneme. In the present work, we used the manually picked time frames for the time alignment step for accurate quantitative analysis.

\subsection{Tongue motion and strain analysis in the subject space}

\def\FigureHeight{94mm}
\begin{figure}[h]
 \center{
 \begin{tabular}{c@{ }c}
   \includegraphics[trim=0mm 0mm 0mm
0mm,clip=true,height=\FigureHeight]{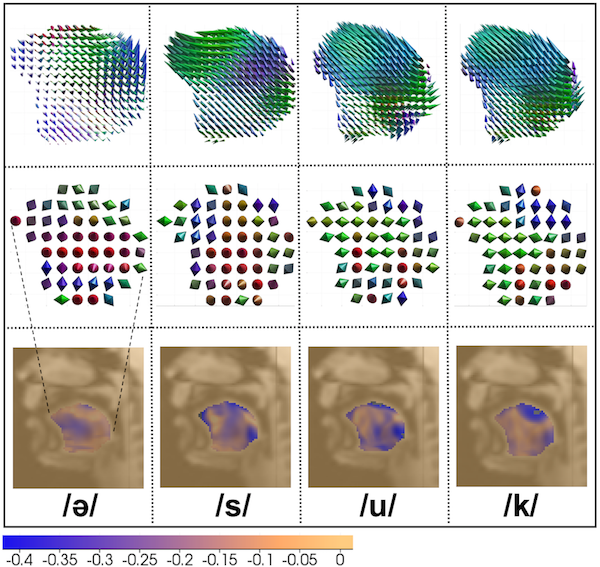} \\
\includegraphics[trim=0mm 0mm 0mm
0mm,clip=true,height=\FigureHeight]{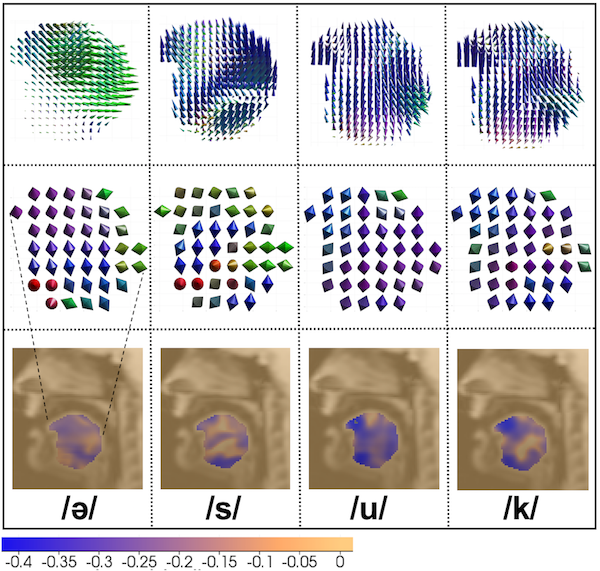} \\
\end{tabular}}
\caption{Subject-specific Lagrangian speech motion fields (first row) are shown with the third Lagrangian strain direction and magnitude (second row) of a healthy subject (top) and a glossectomy patient (bottom) at four representative time frames. The third row shows the fixed anatomic configuration of each subject. The third Lagrangian strain in the second row indicates compression of each tissue point. Note that the cone size and color indicate magnitude and direction (red for left-right, green for front-back, and blue for up-down), and all the analyses are performed in the subject space.}\label{fig:individual}
\end{figure}
%%%%%%%%%%%%%%%%%%%%%%%%%%%%%%%%%%%%%%%%%%%%%%%%%%%%%%%%%%%%%%%%%%%%%

The motion fields, the directions, and magnitudes of the third Lagrangian strain (E3) indicating compression (2D mid-sagittal only) are depicted in Fig.~\ref{fig:individual} for the four time frames for a healthy control and a glossectomy patient. Please note that in this Lagrangian framework, in the course of speech, the tongue configuration appears frozen as shown in Fig.~\ref{fig:individual} (bottom row, cine MR image), but each voxel location depicts the speech movements and associated strain of a fixed tissue element. For the normal control in Fig.~\ref{fig:individual} (left), in the production of /\textschwa/, the genioglossus muscle is compressed slightly more than other regions are as indicated by the blue color. In the productions of /s/ and /u/, the tip, front, and root of the tongue are compressed as indicated by the blue color, while in the production of /k/, the back of the tongue is markedly compressed. Please note that although we have full 3D plus time Lagrangian strains and motion fields, we only show a 2D mid-sagittal slice and therefore the interpretation could be limited. For the glossectomy patient in Fig.~\ref{fig:individual} (right), the motion is different from that of the normal control, in which compressed tongue regions are rather unpredictable due to the tongue resection. For instance, twisting motion pattern is observed in /s/, while most of the tongue regions, the tip, and floor of the tongue are compressed in /u/ and /k/ as reflected in the third Lagrangian strain (E3), respectively, as visually assessed.

%% Table 1 %%%%%%%%%%%%%%%%%%%%%%%%%%%%%%%%%%%%%%%%%%%%%%%%%%%%%%%
\begin{table}[h]
\caption{Lagrangian strains averaged over the whole tongue for normal controls in the atlas space (mean$\pm$SD)}\vspace{0.1cm}
\centering
%\scriptsize{ 
\begin{tabular}{c||c|c|c} \hline
 Lagrangian Strains & E1 & E2 & E3 \\
 \hline \hline
 /\textschwa/  & 0.113$\pm$0.057 & -0.003$\pm$0.005 & -0.089$\pm$0.032  \\ 
 /s/ &  0.183$\pm$0.080 & -0.003$\pm$0.007 & -0.100$\pm$0.035   \\ 
 /u/ &  0.249$\pm$0.120 & 0.008$\pm$0.017 & -0.110$\pm$0.044  \\ 
 /k/ &  0.251$\pm$0.122 & 0.009$\pm$0.015 & -0.114$\pm$0.044   \\
 \hline 
\end{tabular}\label{table:mean_strains}
\end{table}
%%%%%%%%%%%%%%%%%

\def\FigureHeight{100mm}
\begin{figure}[htb]
 \center{
 \begin{tabular}{c@{ }c}
   \includegraphics[trim=0mm 0mm 0mm
0mm,clip=true,height=\FigureHeight]{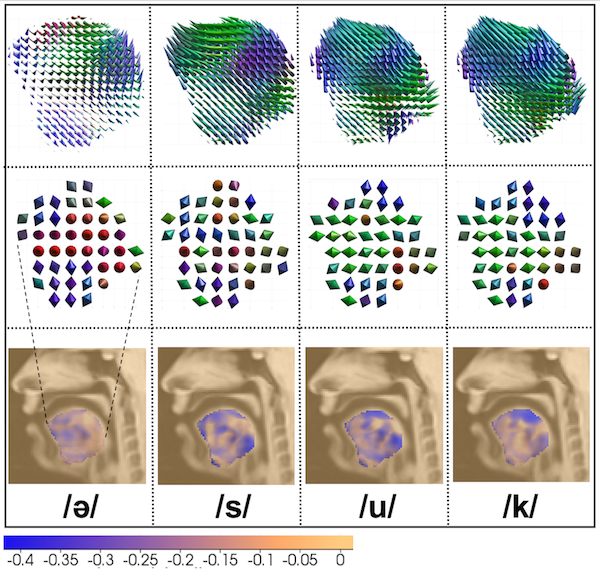} \\
\includegraphics[trim=0mm 0mm 0mm
0mm,clip=true,height=\FigureHeight]{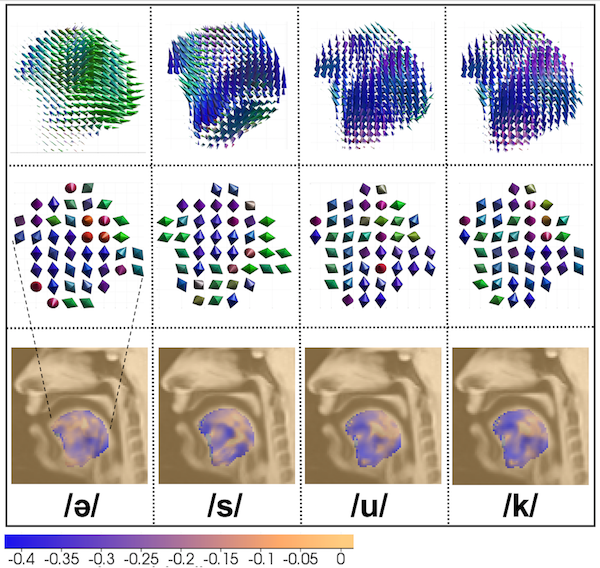} \\
\end{tabular}}
\caption{Lagrangian speech motion fields (first row) in the atlas space are shown with the third Lagrangian strain direction (second row) and magnitude (third row) of a healthy subject (top) and a glossectomy patient (bottom) at the four representative time frames. The third row shows the fixed anatomic configuration of the atlas. The third Lagrangian strain in the second row indicates compression of each tissue point. Note that the cone size and color indicate magnitude and direction (red for left-right, green for front-back, and blue for up-down).}\label{fig:atlas}
\end{figure}
%%%%%%%%%%%%%%%%%%%%%%%%%%%%%%%%%%%%%%%%%%%%%%%%%%%%%%%%%%%%%%%%%%%%%

\subsection{Tongue motion and strain analysis in the atlas space}

\def\FigureHeight{50mm}
\begin{figure}[htb]
 \center{
 \begin{tabular}{c@{ }c@{ }c}
   \includegraphics[trim=0mm 0mm 0mm
0mm,clip=true,height=\FigureHeight]{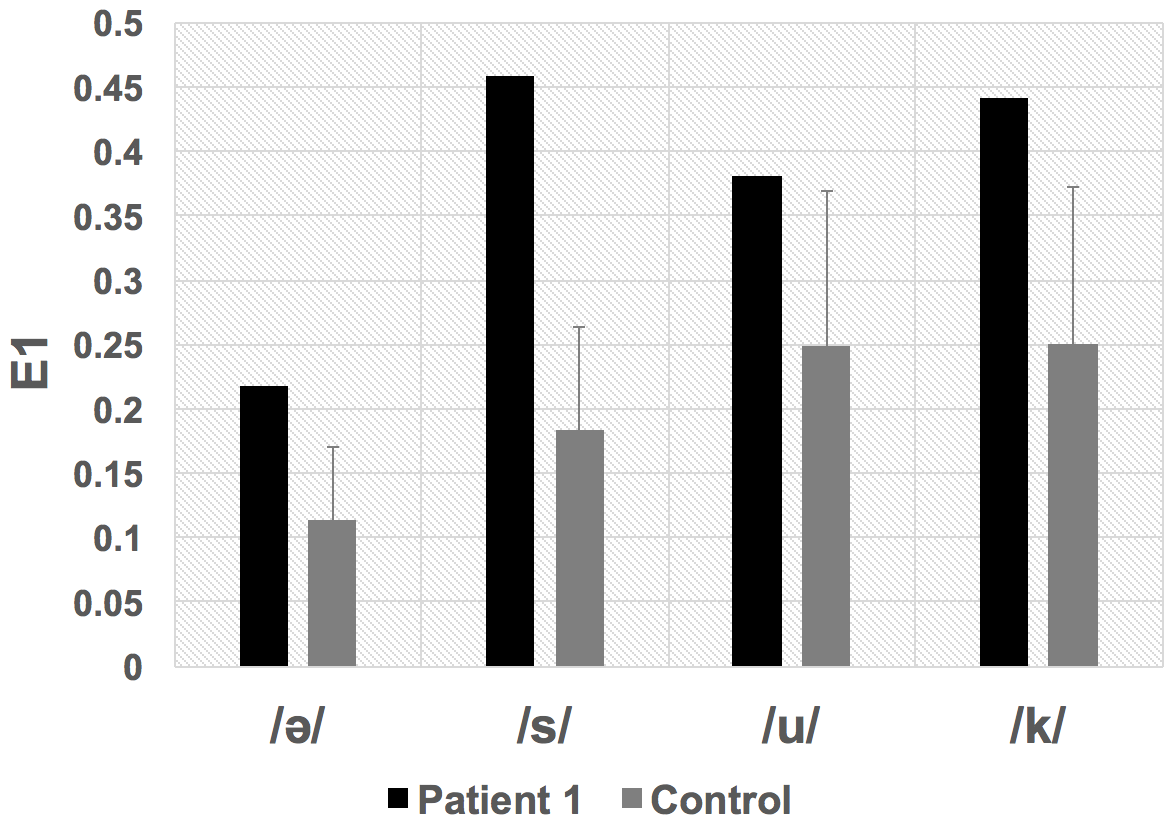}
\includegraphics[trim=0mm 0mm 0mm
0mm,clip=true,height=\FigureHeight]{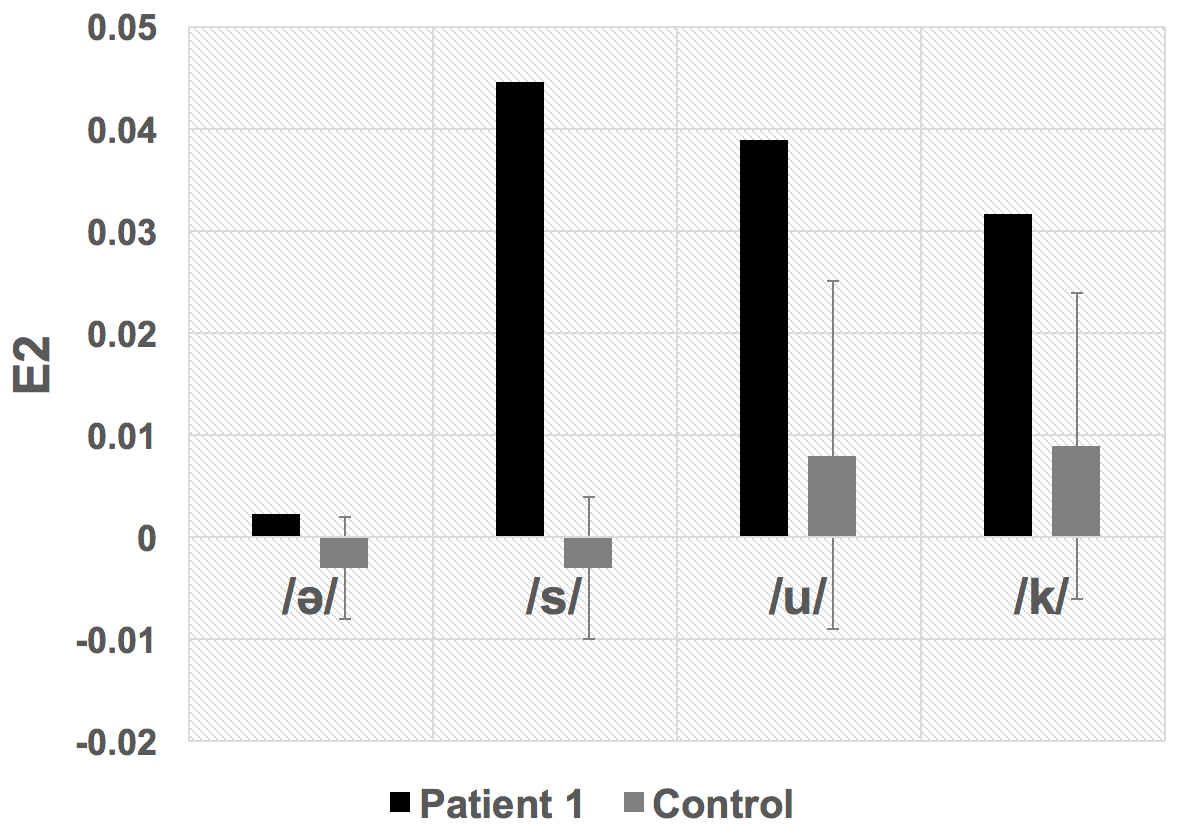} \\
\includegraphics[trim=0mm 0mm 0mm
0mm,clip=true,height=\FigureHeight]{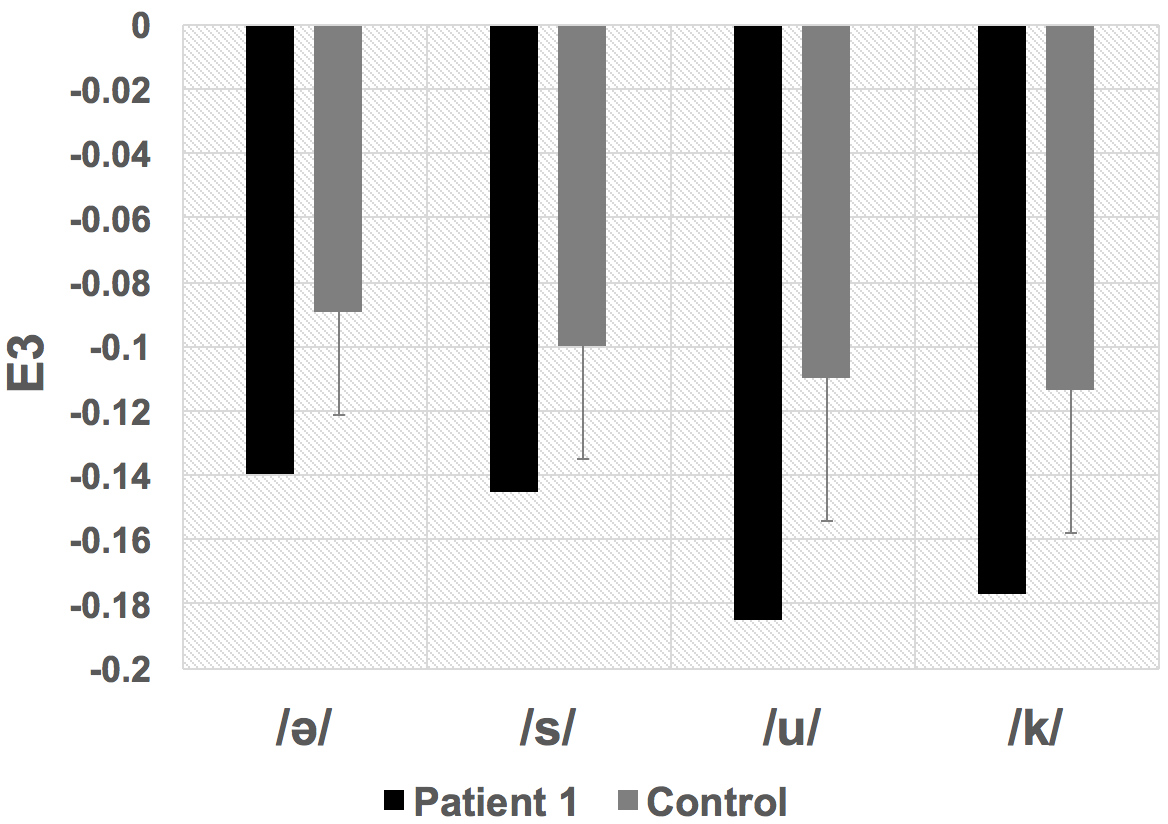} \\
\end{tabular}}
\caption{Plots of the three Lagrangian strain values from the glossectomy patient versus the mean of the three Lagrangian strain values from the normal controls in the atlas space. Please note that the values are derived from the whole tongue. It is shown that the values from the glossectomy patient is higher than the mean of the values from the fourteen normal controls.}\label{fig:strain}
\end{figure}
%%%%%%%%%%%%%%%%%%%%%%%%%%%%%%%%%%%%%%%%%%%%%%%

 Fig.~\ref{fig:atlas} shows the same two subjects used in Fig.~\ref{fig:individual} transformed to the atlas space, where the motion fields, directions, and magnitudes of the third Lagrangian strain (E3) are depicted in the first, second, and third rows, respectively. The directions and patterns of the third Lagrangian strain (as in the second and third rows) and all the Lagrangian strain values between the subject and atlas space are similar and highly correlated (r=0.99, $p$=NS), respectively, indicating that the transformations used in Eq.~(\ref{eq:combination}) preserve the properties of strain in the subject space. Since our proposed Lagrangian framework uses the same material coordinates, it is easy to compare and contrast the differences of strains across subjects. In addition, Table~\ref{table:mean_strains} lists the mean and standard deviation of the Lagrangian strains (E1, E2, and E3) averaged over the whole tongue for all normal controls in the atlas space. Fig.~\ref{fig:strain} depicts the three Lagrangian strain values of a glossectomy patient as in Fig.~\ref{fig:atlas} (right) versus the mean of the normal controls as in Table~\ref{table:mean_strains}. For this glossectomy patient, it is shown that all three Lagrangian strains of the patient are elevated compared to those of the normal controls, suggesting that this patient requires more tissue compression or expansion throughout the whole tongue to produce target sounds.

\subsection{PCA analysis on motion fields}

The PCA was performed on the speech motion data following the transformation of the motion fields (i.e., PVIRA) of each subject to the atlas space based on cine-MRI. For all four time frames, the mean motion fields appear to be reasonable representative motion fields of the classes of each phoneme as shown in Fig.~\ref{fig:pca} when visually assessed. Our PCA results indicate that there are no dominant tongue behaviors in the atlas space as reflected by the variance of different PCs, where the first three PCs for all the time frames accounted for less than 45$\%$ as shown in Table~\ref{table:PC_loading}. This is partly because the PCA analysis done in the atlas space captures the differences and variability in speech motion after the size and shape differences of the tongue across subjects were corrected to minimize the effect of anatomical differences. The PCA results thus capture subtle yet objective motion differences in lingual motion across speakers. First, in the production of /\textschwa/, the first PC, which accounts for 19.53\%, seems to capture forward and backward motion in the front part of the tongue, while the second PC, which accounts for 14.03\%, roughly captures downward and backward motion and the third PC, which accounts for 9.47\%, captures the rotating motion. Second, in the production of /s/, the first PC, which accounts for 16.51\%, appears to differentiate between apical /s/ (-1$\sigma$) and laminal /s/ (+1$\sigma$), in which apical /s/ is produced with the tip of the tongue, while laminal /s/ is produced with the blade of the tongue. The second and third PCs, which account for 13.08\% and 9.95\%, capture forward and upward for the second PC, and upward and forward motion for the third PC in the lower half of the tongue, respectively. Third, in the productions of /u/ and /k/, the first PC, which accounts for 17.68\% and 17.95\%, captures the different use of the tongue tip and blade as in /s/. The second PC, which accounts for 12.21\% and 12.42\%, explains the use of the front and back of the tongue, while the third PC, which accounts for 11.58\% and 11.05\%, contrasts upward (-1$\sigma$) and forward (+1$\sigma$) motion for /u/ and upward (-1$\sigma$) and forward (+1$\sigma$) motion for /k/, respectively. 

\def\FigureHeight{61mm}
\begin{figure}[htb]
 \center{
 \begin{tabular}{c@{ }c@{ }c}
   \includegraphics[trim=0mm 0mm 0mm
0mm,clip=true,height=\FigureHeight]{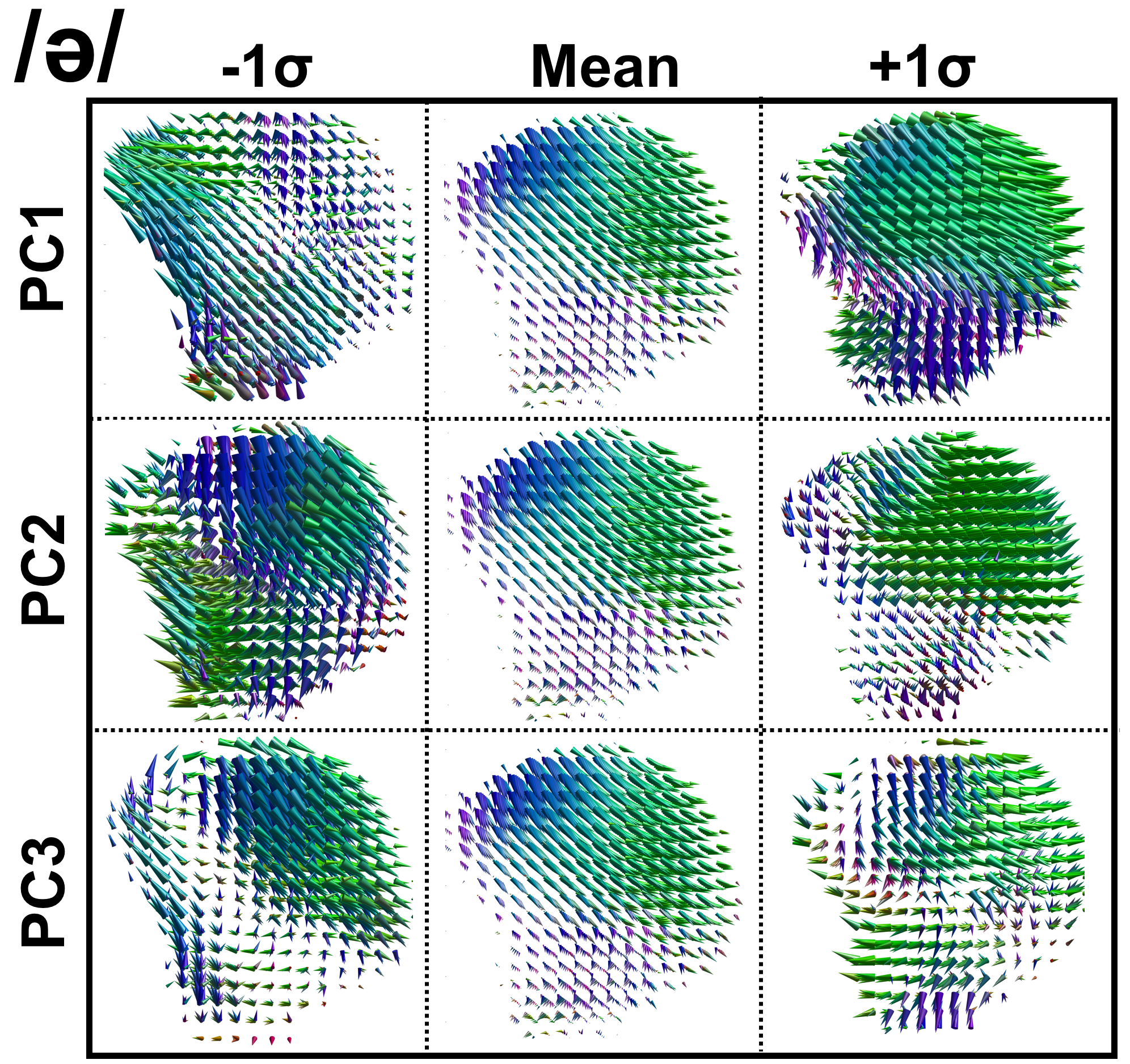}
\includegraphics[trim=0mm 0mm 0mm
0mm,clip=true,height=\FigureHeight]{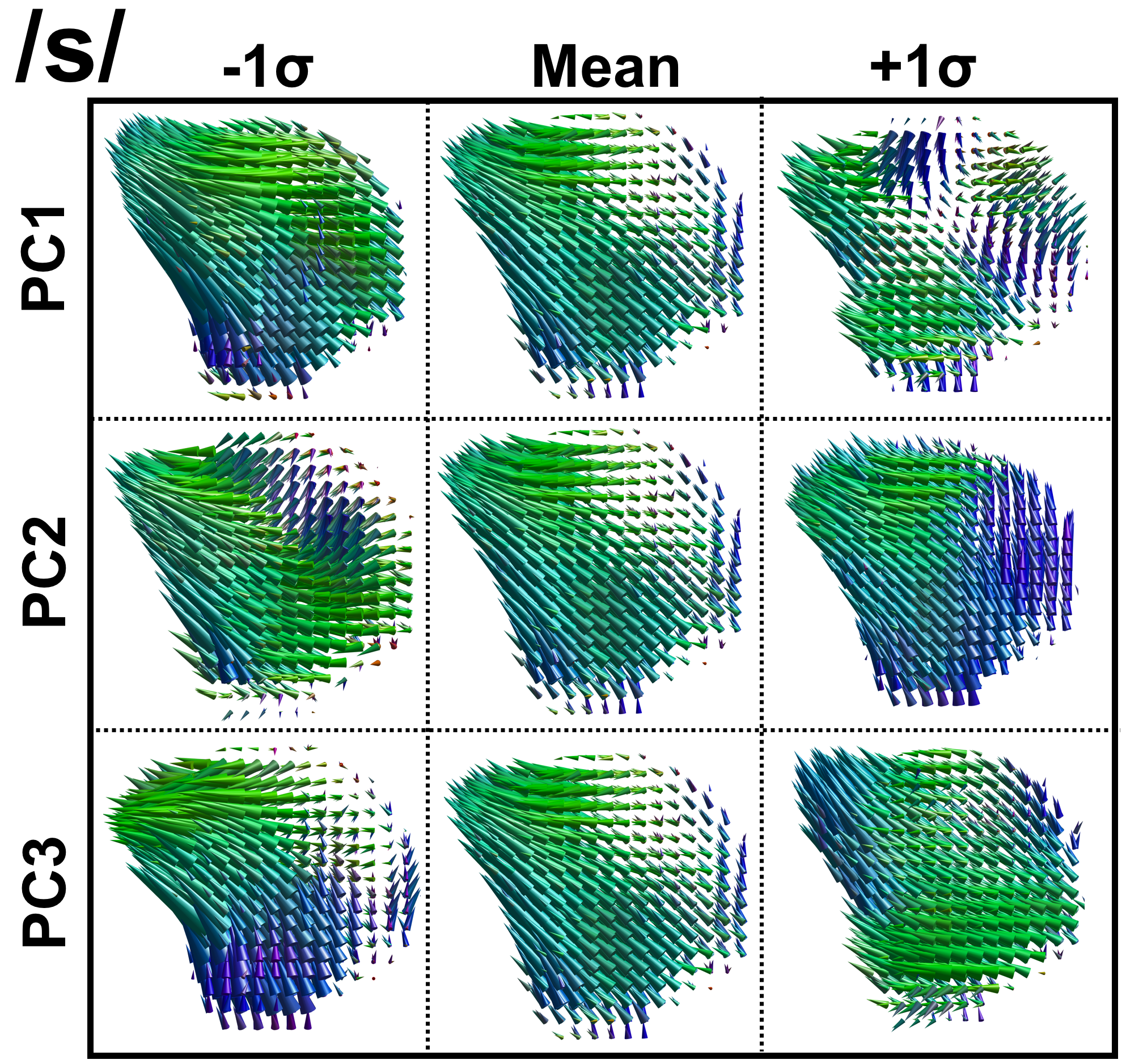} \\
\includegraphics[trim=0mm 0mm 0mm
0mm,clip=true,height=\FigureHeight]{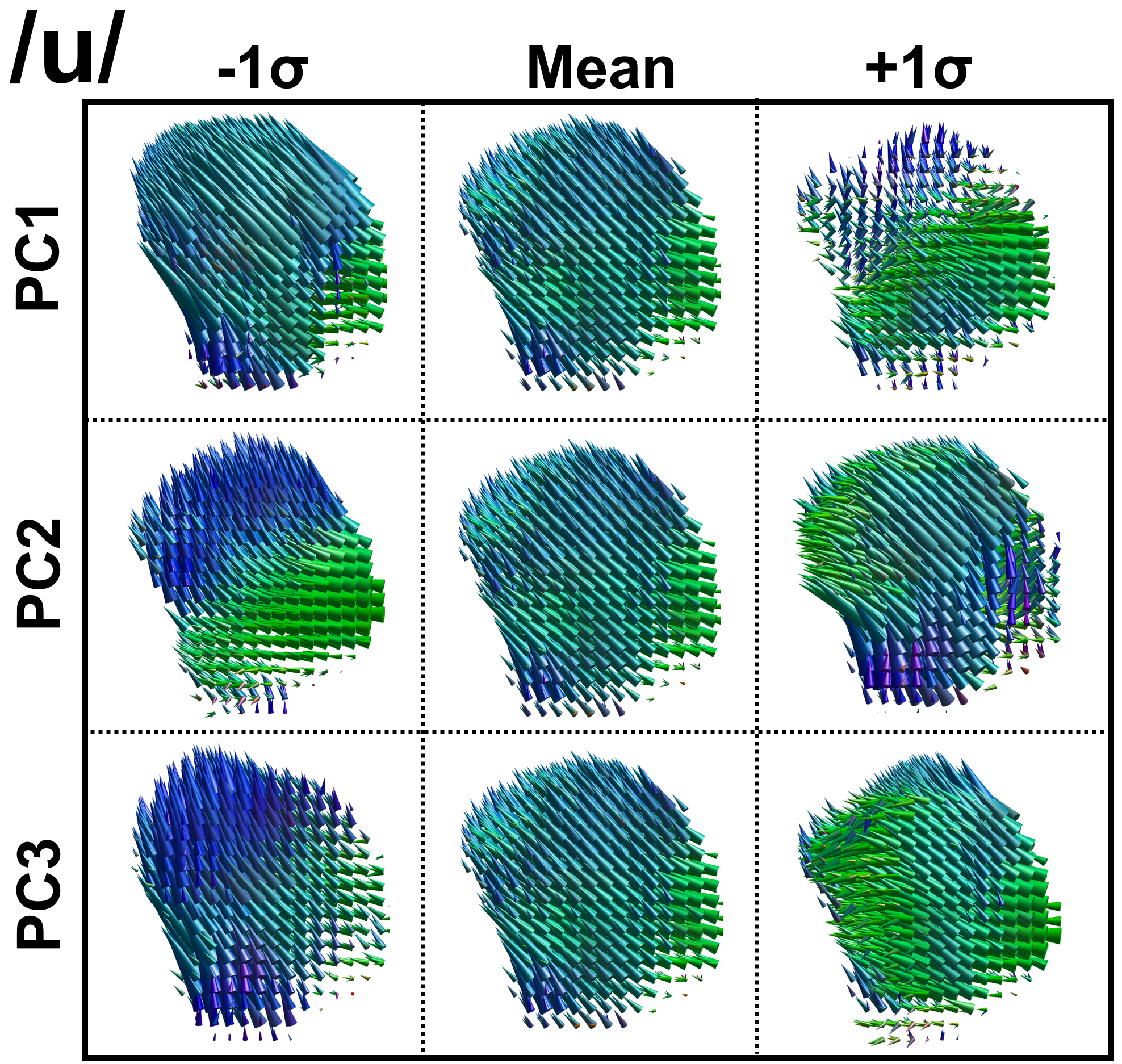} 
\includegraphics[trim=0mm 0mm 0mm
0mm,clip=true,height=\FigureHeight]{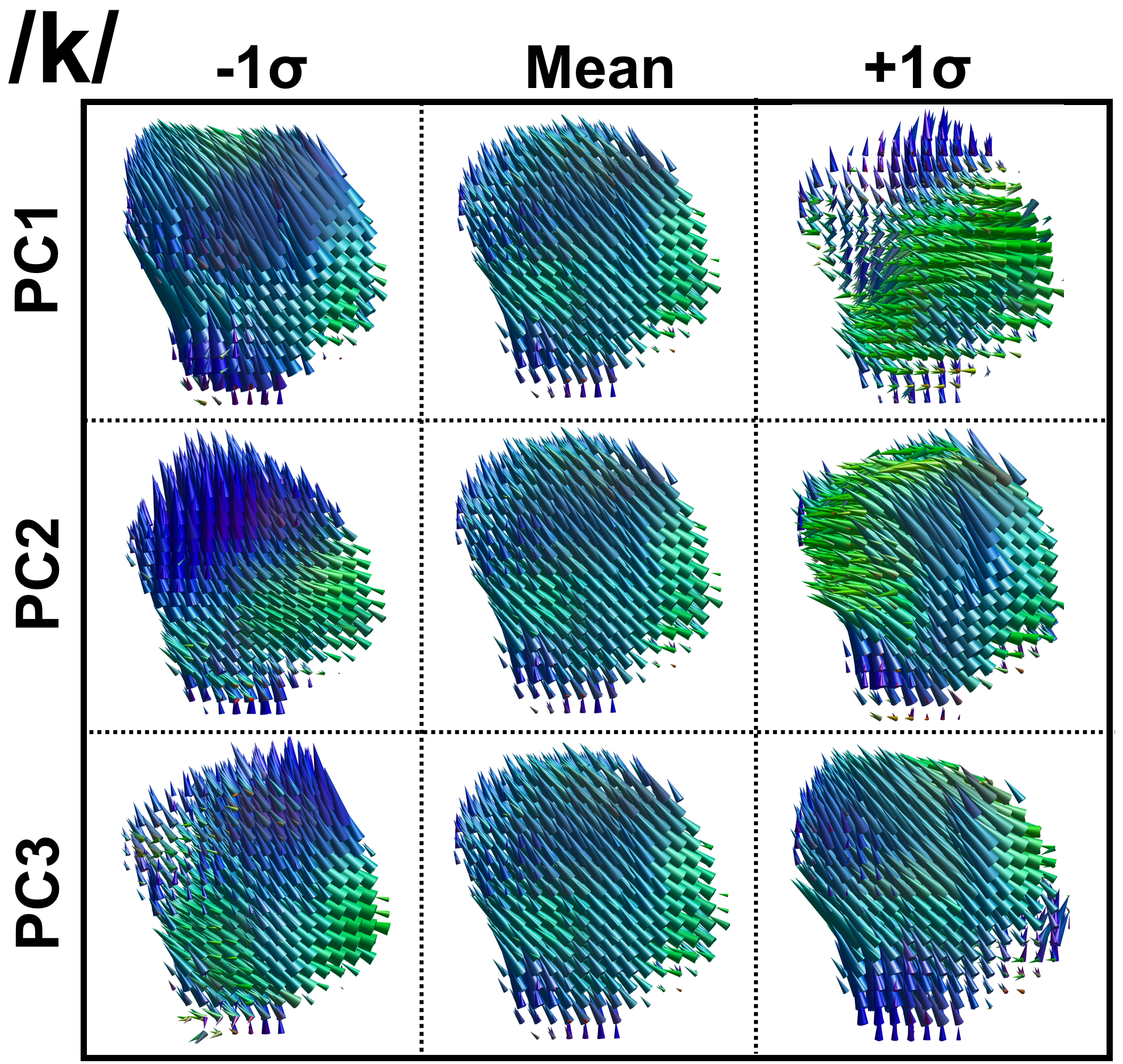} \\
\end{tabular}}
\caption{The three primary PCs of variance of the motion fields from ``\textschwa-suk.''}\label{fig:pca}
\end{figure}
%%%%%%%%%%%%%%%%%%%%%%%%%%%%%%%%%%%%%%%%%%%%%%%

%% Table 1 %%%%%%%%%%%%%%%%%%%%%%%%%%%%%%%%%%%%%%%%%%%%%%%%%%%%%%%
\begin{table}[h]
\centering
\caption{PC loadings for the four time frames (\%)}
\begin{tabular}{c|c|c|c} \hline
 PC & PC1 & PC2 & PC3   \\
 \hline \hline
 /\textschwa/ & 19.53 & 14.03 & 9.47  \\
 /s/ & 16.51 & 13.08 & 9.95  \\
 /u/ & 17.68 & 12.21 & 11.58  \\
 /k/ & 17.95 & 12.42 & 11.05 \\
  \hline 
\end{tabular}\label{table:PC_loading}
\end{table}
%%%%%%%%%%%%%%%%%

\section{Discussion}

\subsection{Summary of Results}

In this work, we present a novel approach to visualizing and analyzing tongue motion during speech by constructing a statistical multimodal atlas of 4D tongue motion within a Lagrangian framework. Integrative models of tongue anatomy and physiology using multimodal tongue imaging play an important role in characterizing tissue function and properties. Our atlas framework based on a normal population is an advantage to studying the mechanisms of speech production in normal and diseased populations including tongue cancer and brain disorders such as amyotrophic lateral sclerosis (ALS) or stroke. We report several new findings that would have been difficult to obtain with existing methods. First, unlike other approaches (e.g.,~\cite{woo_cmbbe_2016,eulerian_atlas}), within a Lagrangian framework, we show for the first time the directions and magnitudes of the Lagrangian strain and internal motion patterns on a subject-by-subject basis and further in the atlas space. In the subject-specific material coordinate system, it is possible to compare internal motion and strain patterns across different sounds, while in the atlas material coordinate system, it is possible to compare the internal motion and strain patterns across subjects in an objective and quantitative manner. In this way, we can create a motion map that is not biased by a specific individual's anatomical and functional features. Second, this approach allows us to capture motion variability by providing a fixed coordinate space for motion analysis using PCA. For instance, PCA analysis on /s/ production captures the difference between apical and laminal /s/. Third, in studying patients, our approach can be used to compare abnormal behaviors in relation to normal motor variation using internal motion fields and strain in the atlas space. This atlas space allows us to compare motion information ranging from voxel level to muscle level. As shown in Table~\ref{table:mean_strains}, we established the average values of the strain in the whole tongue with which to compare against strain values of patient data. 

\subsection{A comparison of Lagrangian and Eulerian frameworks}

In the present work, we chose to use a Lagrangian configuration rooted in the material coordinates. Since the material coordinate system does not experience any deformation in every following time frame and all quantities are mapped back to this static configuration, the process significantly simplifies comparison of the strain patterns during speech across subjects. Therefore, a ``motionless'' concept~\cite{motionless} is established in the context that all motion is converted to the form of changing value of specific variables.

Another strategy is to use similar processes applied to an Eulerian configuration rooted in the spatial coordinates, such as reported in~\cite{eulerian_atlas}. Such a ``moving atlas'' allows for visualization of the moving tongue anatomy parallel to the cine-MRI in the form of changing motion fields during speech. The strength of such an Eulerian atlas is that it reflects the real deforming properties of the tongue shape and motion fields, since the shape changes are also accounted for in the coordinate changes, which can then be directly applied to the deformation and comparison of cine-MRI. Besides, strain and other properties at a specific time instance can be directly computed without additional deformation. However, temporal analysis of any changing variable may become more difficult as coordinates of the whole tongue keep changing, and inverting tracking of any fixed tissue point is not feasible.

\subsection{Time alignment of speech movements across subjects}

In the present work, we focused on the analysis of the four time frames by manually picking the four time frames for each subject. In order to study the whole phrase, however, it is necessary to accurately align the speech task across subjects to build a temporally aligned 4D atlas. This is a challenging task as there is a high inter-subject variability in speech movements even after training each subject to speak to a metronome. This could be more problematic when building 4D atlases using real-time MRI such as~\cite{Naranayan2004,fu2016speech}, since it is difficult to control an individual speaker's tempo as opposed to the data collection strategy using repeated utterances to acquire cine-MRI with a 1-second duration as described in this work. One can tackle this problem using either motion quantities derived from tagged-MRI such as strain and the mean of the magnitude of the motion field as in Eq. (\ref{eq:mean_deform}) or motion quantities derived from cine-MRI~\cite{woo_cmbbe_2016,realtime_fu}. These quantities can serve as a motion descriptor to find temporal correspondences across subjects. Additionally, if speech acoustic samples that are synchronized with cine-MRI data are available, then one may consider using speech acoustic samples to find temporal correspondences across subjects using a dynamic time warping approach~\cite{dtw}, and then apply the alignment to the imaging data.   

\subsection{Using Lagrangian strain as a marker of muscle activation}

Internal tongue motion patterns and associated strain measurements provide a link between muscle activation and tongue surface shape. Although strain measurements indicating tissue compression and expansion can be used as a useful surrogate (or biomarker) for muscle activation, their relationship to actual muscle activation is complex and variable from one subject to another. The challenge is mainly because a large number of muscles are inter-digitated and activate in different patterns to create a deformation, leading to complex strain directions that are difficult to quantitatively and visually assess. Our atlas framework could provide new insights into the understanding of tongue muscle coordination and muscle activation by providing average motion fields and principal directions and magnitude of the strain. More studies, however, are needed using electromyography (EMG)~\cite{emg} or biomechanical simulations~\cite{biomechanics} to investigate this along with our framework to a great extent.

\subsection{Using diffeomorphic registration to create the 4D atlas during speech}
Our 4D atlas approach relies on the accurate registration of tongue regions and associated motion fields to a common template to localize motion changes during speech. Registration may work better for normal controls than for patient data, especially in glossectomy patients, depending on the regional homogeneity in the tongue that is used to find the correspondences. Some patients have heterogeneous tissue types (e.g., glossectomies) or smaller tongues depending on different disease states (e.g., ALS). Therefore, it is necessary to perform some specialized preprocessing to reduce registration error and potential bias. For example, one can segment the tongue region and use the corresponding segmented regions to provide initial anatomical landmarks for the registration method that help localize and emphasize the region of interest.

\subsection{Interfacing between motion fields and anatomy in the 4D atlas}
Measuring internal motion patterns is only the first step towards explaining the intramural mechanics of the human tongue in association with physiological deformations during speech. Next one needs to put the analysis including strain measurements in the context of the tongue's muscular anatomy derived from diffusion MRI, structural MRI~\cite{strain_dti} or structural atlases~\cite{hatlas} to investigate muscle activation along with muscle anatomy such as muscle shortening. Multimodal registration such as~\cite{multimodal_reg} could be used to interface between motion fields and anatomy depending on the imaging modalities being considered. In this way, our atlas framework could allow us to compare motion information at different resolution levels ranging from voxel level to individual muscle level to muscle group level (e.g., functional units~\cite{functional_units}). In addition, linking internal motion fields and associated strain patterns to muscle anatomy in the subject as well as atlas space has a potential to shed light on the ``functional organization'' such as functional units~\cite{functional_units} of the tongue during speech. 

\subsection{Tracking tissue points using phase-based versus intensity-based registration approach}
To track each point of the tongue from tagged-MRI, we used PVIRA in this work. PVIRA works on the harmonic phase volumes extracted from tagged-MRI, while iLogDemons works directly on the intensity volumes of the tagged data. Since harmonic phase is a physical property directly related to actual tissue location, direct matching of phase values is more reliable than matching of intensity values that are prone to tag fading and noise. In the original HARP~\cite{harp}, it has been demonstrated that the use of phase normally results in tracking errors less than a third of the pixel resolution, while matching of intensity is more likely to fail when tags fade at later time frames. Therefore, using PVIRA instead of iLogDemons is a natural choice to reduce the impact of intensity change over all time frames. In related work, tracking internal points of the tongue in a 2D mid-sagittal slice using HARP from tagged-MRI has showed superior performance to various intensity-based registration approaches from cine-MRI~\cite{comparison_Woo}.

\subsection{Computing statistics on diffeomorphisms to characterize individual subject's vocal tract parameters}
Since its inception of ``computational vocal tract anatomy''~\cite{hatlas}, efforts have been made to understand and model not just the anatomy itself but the anatomical changes of the tongue over time and its variations across a population~\cite{hatlas_stone, woo_cmbbe_2016}. Although this first effort only accounts for the tongue, we can expand the reference anatomic configuration and the motion fields to include the whole vocal tract. In addition, although registration of each subject with the atlas has the effect of warping the vocal tract that would not be expected to preserve articulatory-acoustic relations, the diffeomorphisms used to warp each individual subject to the atlas encode information on parameters that reveal characteristics of each individual's vocal tract such as vocal tract area function. This is similar to the approaches used in computational anatomy~\cite{miller}. 

\section{Conclusions and Future Directions}

In this work, a statistical multimodal atlas of 4D tongue motion from both cine- and tagged-MRI has been successfully constructed using healthy subjects for the speech task, ``\textschwa-suk.'' To our knowledge, this atlas is the first of its kind, thus opening new vistas to study the relationship between structural and functional properties of the tongue during speech. In our future work, we will further investigate classification and statistical techniques for categorizing groups of subjects. For instance, in addition to PCA, deep learning based approaches~\cite{deep_learning} will be carried out to perform regression and classification using the multidimensional features including displacements, strain, or muscle mechanics derived multimodal imaging data to differentiate between the normal motion pattern from our atlas and pathologic motion patterns from patient populations. In addition, our approach can be broadly applied to analyze how tongue function for speech is limited by abnormal internal motion and strain in a variety of patient groups such as patients who have undergone treatment for cancer or other diseases such as aphasia or impaired language development caused by brain injury.

\section{Acknowledgements} 
This research was supported in part by NIH R00DC012575, R01DC014717, R01CA133015, S10OD011928, NSF PHY1504804, and ECOR ISF funding. We thank Euna Lee for proofreading the text.

\end{document}